\documentclass[lettersize,journal]{IEEEtran}
\usepackage{amsmath,amsfonts}
\usepackage{array}
\usepackage[caption=false,font=normalsize,labelfont=sf,textfont=sf]{subfig}
\usepackage{amsthm,amsmath,amssymb}
\usepackage{algorithm}

\usepackage{ulem}

\usepackage{makecell}
\usepackage{graphicx}
\usepackage{dsfont} 
\usepackage{amsmath}
\usepackage{amssymb}
\usepackage{booktabs}
\usepackage{bbm}

\usepackage{epsfig}
\usepackage{makecell,rotating}
\usepackage{array}
\usepackage{colortbl}
\usepackage{xcolor}
\usepackage{multirow}
\usepackage{algpseudocode}

\usepackage{mathrsfs}

\usepackage{textcomp}
\usepackage{booktabs}
\usepackage{stfloats}
\usepackage{url}
\usepackage{verbatim}
\usepackage{graphicx}
\usepackage{cite}
\usepackage{amssymb}

\usepackage[pagebackref,breaklinks,colorlinks]{hyperref}
\usepackage{fancyvrb}
\usepackage{url}

\begin{document}

\title{Reconciling Semantic Controllability and Diversity for Remote Sensing Image Synthesis with\\ Hybrid Semantic Embedding}

\author{Junde~Liu,
        Danpei~Zhao*,~\IEEEmembership{Member,~IEEE,}
        Bo~Yuan,
        Wentao~Li,
        Tian~Li
\thanks{This work was supported by the National Natural Science Foundation of China under Grant 62271018. (Corresponding author: Danpei Zhao.)}
\thanks{Danpei Zhao, Bo Yuan and Tian Li are with the Image Processing Center, School of Astronautics,  Beihang University, Beijing 102206, China, and also with Tianmushan Laboratory, Hangzhou 311115, China (e-mail: zhaodanpei@buaa.edu.cn, yuanbobuaa@buaa.edu.cn, lit@buaa.edu.cn).}
\thanks{Junde Liu and Wentao Li are with the Image Processing Center, School of Astronautics, Beihang University, Beijing 102206, China (e-mail: \mbox{jundeliu@buaa.edu.cn}, canoe@buaa.edu.cn).}

\thanks{Manuscript received XX, 2024; revised XX.}}

\markboth{Journal of \LaTeX\ Class Files,~Vol.~XX, No.~XX, XXXX}
{Shell \MakeLowercase{\textit{et al.}}: A Sample Article Using IEEEtran.cls for IEEE Journals}

\maketitle

\begin{abstract}
Significant advancements have been made in semantic image synthesis in remote sensing. However, existing methods still face formidable challenges in balancing semantic controllability and diversity. In this paper, we present a Hybrid Semantic Embedding Guided Generative Adversarial Network (HySEGGAN) for controllable and efficient remote sensing image synthesis. Specifically, HySEGGAN leverages hierarchical information from a single source. Motivated by feature description, we propose a hybrid semantic Embedding method, that coordinates fine-grained local semantic layouts to characterize the geometric structure of remote sensing objects without extra information. Besides, a Semantic Refinement Network (SRN) is introduced, incorporating a novel loss function to ensure fine-grained semantic feedback. The proposed approach mitigates semantic confusion and prevents geometric pattern collapse. Experimental results indicate that the method strikes an excellent balance between semantic controllability and diversity. Furthermore, HySEGGAN significantly improves the quality of synthesized images and achieves state-of-the-art performance as a data augmentation technique across multiple datasets for downstream tasks.

\end{abstract}


\section{Introduction}
\IEEEPARstart{S}{emantic} image synthesis in remote sensing aims to generate semantic-controllable synthetic images based on semantic masks.
This form of conditional image synthesis is crucial for applications like semantic image editing~\cite{luo2023siedob, ling2021editgan, ntavelis2020sesame}, image segmentation~\cite{Tasar2019IncrementalLF, BAFFT, yuan2024csssurvey, yuan2022birds, zhao2023inherit, yuan2024learning}, and data enhancement~\cite{triantafyllidou2020low, chen2023deep, pernuvs2024fice}. In these applications, semantic controllability, diversity and extensibility of the synthesized images are essential.
\begin{figure}[!t]
\centering
\includegraphics[width=3.51in]{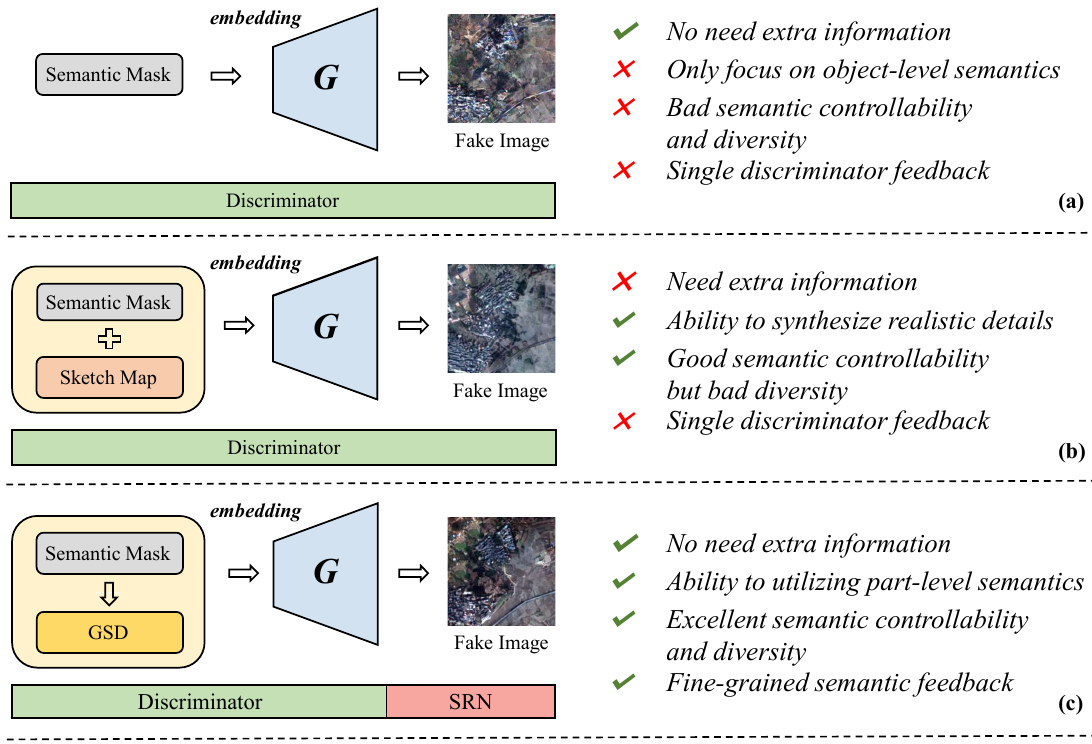}
\caption{Schematic diagram of the proposed method. (a) Traditional semantic image synthesis method~\cite{park2019spade}, which exhibits poor performance. (b) Introducing additional information for fused semantic embedding. For instance,~\cite{tan2023inade} enhances synthesized semantic features with sketches, yet it still encounters issues like class confusion in the results. (c) Proposed hybrid semantic Embedding and fine-grained semantic feedback mechanism. Our approach ensures semantic consistency and extensibility, minimizing the sacrifice of diversity. The results demonstrate that our strategy effectively enhances the quality of fine-grained synthesized images for remote sensing image synthesis tasks.}
\label{fig-motivation}
\end{figure}
Semantic controllability in image synthesis involves two key aspects: semantic consistency and generation quality. Semantic consistency refers to the alignment between synthesized images and masks, while generation quality pertains to the visual realism, including factors like color, texture, and accurate depiction of ground features. Diversity, on the other hand, refers to the model's ability to synthesize multiple visually distinct images from a single mask~\cite{dong2017semantic,wang2018p2phd,park2019spade,zhu2020sean}.
\par 
Many existing approaches focus on global diversity rather than semantic diversity. They leverage the variational autoencoder (VAE)~\cite{kingma2013auto} architecture to constrain one-one generation. Regarding semantic controllability, numerous recent studies~\cite{dong2017semantic, isola2017p2p, wang2018p2phd, park2019spade, zhu2020sean, tang2020local, tan2023inade, tan2022clade} emphasize the quality aspect of image generation. These methods transform semantic masks into general image-to-image frameworks by either feeding them directly into an encoder-decoder network~\cite{isola2017p2p, wang2018p2phd} or using spatial adaptive normalization~\cite{park2019spade} and its variants~\cite{tan2022clade, sushko2022oasis}. However, as shown in Fig.~\ref{fig-motivation}(a), these approaches focus mainly on global semantics, overlooking the finer-level semantic layout. This oversight leads to decreased semantic consistency and limited fine-grained controllability. Some studies~\cite{tan2023inade, tang2023ecgan, shi2022resail} attempt to constrain synthesis results by incorporating new semantic cues, like sketch maps, but still fail to provide fine-grained semantic feedback. These approaches are generally depicted in Fig.~\ref{fig-motivation}(b). Although recent methods, such as~\cite{lv2024place} and~\cite{xue2023freestyle}, strive to balance these aspects, the improvements in synthesized image quality are still inadequate.
\par 
Since the aforementioned methods are primarily designed for natural images or everyday objects, their performance diminishes significantly when applied to remote sensing images. Remote sensing targets exhibit substantial semantic ambiguity, complex occlusion and irregular spatial distribution within instances. Semantic ambiguity denotes the extensive feature overlap between different categories of remote sensing objects (e.g., grass and forest) at the semantic level. Complex mutual occlusion describes the spatial overlap of remote sensing objects, either within the same category or across different categories, creating challenges for the training process. Irregular spatial distribution within instances pertains to the varying shapes and geometric arrangements of remote sensing objects, such as buildings. These inherent features lead to issues like synthetic semantic confusion and geometric pattern collapse, resulting in poor semantic controllability and diversity. Consequently, addressing these issues is essential for achieving a proper balance between controllability and diversity. The balance between these aspects can be directly evaluated using downstream tasks, such as semantic segmentation. If the synthesized images are of adequate quality, using them for data enhancement can significantly improve the performance of downstream tasks.
Therefore, incorporating improved downstream task metrics is crucial when evaluating synthesized remote sensing images, which have often been overlooked in previous studies~\cite{dong2017semantic,isola2017p2p,wang2018p2phd,park2019spade,zhu2020sean,tang2020local,tan2022clade,tan2023inade,xue2023freestyle,lv2024place}.
Derived from by CSEBGAN~\cite{wang2023csebgan}, this paper introduces the concept of extensibility. This refers to the effectiveness of synthesized images in enhancing downstream tasks like image segmentation, and establishes a balance between semantic controllability and diversity. To address these challenges, we propose a novel semantic image synthesis method named HySEGGAN (Hybrid Semantic Embedding Guided Generative Adversarial Network), which utilizes hybrid semantic embeddings for remote sensing image synthesis. Fig.~\ref{fig-motivation}(c) illustrates the proposed method. Our approach achieves competitive fidelity and fine-grained diversity, substantially enhancing performance in downstream tasks.

\par
HySEGGAN revolutionizes traditional GAN frameworks by resolving the issue where discriminators must simultaneously evaluate image fidelity and semantic consistency. This dual assessment increases learning complexity and can impede the generator's ability to receive fine-grained semantic feedback. HySEGGAN introduces three key components: (1) Geometric-informed Spatial Descriptor (GSD), (2) Hybrid Semantic embedding Guided Network (HSGNet) and (3) Semantic Refinement Network (SRN). HSGNet enhances diversity and fine-grained controllability by converting semantic masks into GSDs. These descriptors are then embedded as distinct semantic representations. By emphasizing the part-level layout of objects instead of relying solely on global semantics, this method enhances both controllability and diversity. Specifically, HSGNet uses parameter-free attention and convolution mechanisms, delivering superior results with fewer parameters. Furthermore, to tackle semantic confusion, we integrate a novel Semantic Refinement Network into the traditional training scheme and design a novel loss function. During training, the discriminator provides global fidelity feedback to the generator, while the semantic refinement network offers local, fine-grained semantic feedback to maintain semantic mask consistency.
\par 
Experimental results on two very different remote sensing image datasets demonstrate the effectiveness of the proposed method. Compared with other models, HySEGGAN achieves an excellent balance of semantic controllability and diversity. To verify the robustness of the model in applications, we further evaluate its performance in specific classes.
\par
In summary, the contributions of this paper are summarized as follows.
\begin{itemize}
\item[$\bullet$] We propose a remote sensing image synthesis method based on hybrid semantic embedding that balances controllability and diversity, achieving high semantic stability and plasticity in the synthesized images.
\item[$\bullet$] From the perspective of feature description and semantic feature modelling, three modules are designed to address the problems of pattern collapse and semantic obfuscation without any additional supervisory information.
\item[$\bullet$] Comprehensive test results on the GID-15 and ISPRS datasets validate the effectiveness of the proposed model. Moreover, the proposed approach can effectively improve the performance of the segmentation task without requiring additional annotations.
\end{itemize}

\section{Related Work}
\subsection{Semantic Image Synthesis }
Semantic image synthesis focuses on synthesizing realistic images from given semantic masks. The field was first defined by Pix2Pix~\cite{isola2017p2p}, proposed by Phillip Isola et al. in 2017, which tackled the general image-to-image translation problem. Pix2PixHD~\cite{wang2018p2phd} extended this work by enhancing image resolution and improving realism and stability in high-resolution training through coarse-to-fine generators and multi-scale discriminators. Park T et al. introduced GauGAN~\cite{park2019spade}, which replaced semantic labels with the spatially-adaptive normalization layer (SPADE~\cite{park2019spade}). This innovation encoded semantic information directly into the network, mitigating semantic confusion caused by deep convolutional stacking. Subsequent models like SEAN~\cite{zhu2020sean}, RESAIL~\cite{shi2022resail} and SAFM~\cite{lv2022safm} further refined this approach, optimizing the scale and offset learning to achieve superior generative results.
\par 
Subsequent methods incorporated diversity into considerations. GroupDNet~\cite{zhu2020groupd} achieved semantic-level diversity using group convolution. Subsequently, CC-FPSE~\cite{shaham2021cc-fpse} introduced conditional convolution kernel parameters guided by semantic layout prediction. It also utilized a feature pyramid semantic embedding discriminator to enhance image detail quality and semantic alignment. SCGAN~\cite{jiang2021scgan} learned a semantic vector to parameterize both the conditional convolution kernel and normalization parameters. LGGAN~\cite{tang2020lggan} employed both local class-specific and global image-level GANs to independently learn the appearance distributions for each object class and the entire image. INADE~\cite{tan2023inade} modeled parameters for each semantic class independently, assuming a continuous distribution and controlled by novel scaling and offset adjustments of the synthesized image. OASIS~\cite{sushko2022oasis}, by contrast, designed a discriminator based on segmentation networks to offer more effective feedback to the generator, yielding semantically aligned images of higher fidelity. Recently, several other methods, such as BicycleGAN~\cite{meshry2021bicyclegan}, LCGAN~\cite{tang2020lcgan}, SC-GAN~\cite{wang2021sc-gan} and ECGAN~\cite{tang2023ecgan}, explore structural and shape information in semantic mask to improve image quality. \par
Despite significant progress in semantic image synthesis, existing methods continue to struggle with limited quality and diversity, mainly due to the representational constraints of the semantic layout and lack of geometric information. Unlike most methods, our approach extends the semantic information through feature descriptions, which enhances the representation of the model.

\subsection{Semantic Embedding Methods in Image Synthesis }
Image embedding has proven to be an effective strategy for reducing the cost of image synthesis collection through the use of existing data~\cite{tang2020lcgan,jiang2021scgan}. SPADE~\cite{park2019spade} first normalizes the network's activation values to a distribution with mean 0 and variance 1 using a parameter-free method, then remaps them to a new mean and variance. 
Subsequent work~\cite{zhu2020sean,shi2022resail} has explored how control parameters from the style mapping can be merged with those from the semantic label.
CLADE~\cite{tan2022clade} departs from GauGAN's design, learning control parameters from the semantic label map using a convolutional network. It instead directly maintains control parameters for each semantic category at every layer, embedding them as learnable weights in the generative network. Several subsequent approaches have explored a new paradigm in semantic embedding~\cite{tang2023ecgan,sushko2022oasis,tan2023inade}, showing promising results.
\par 
While most existing methods combine control parameters from various sources, they often neglect the geometric information inherent in part-level semantic regions. To address this limitation, this paper presents a hybrid semantic embedding approach for guiding image synthesis, enabling the generator to incorporate richer a priori information. This method comprehensively captures the characteristics of specific semantic objects and produces semantically controllable parameters.
\subsection{Remote Sensing Image Synthesis}
In recent years, semantic image synthesis has garnered significant attention~\cite{tang2023ecgan,tang2020lggan,wang2021sc-gan,tan2023inade,lv2022safm}, however, limited research has focused on the field of remote sensing. Simply adapting advanced natural image synthesis techniques for remote sensing tasks often proves ineffective. The recent method, CSEBGAN~\cite{wang2023csebgan}, for remote sensing image synthesis emphasizes semantic parameter modeling but neglects crucial elements like feature modulation and local semantic consistency. Consequently, this results in limited controllability and a decline in the quality of the synthesized images. Moreover, many other approaches fail to consider the unique spatial and semantic characteristics of remote sensing objects, leading to performance that remains suboptimal.
\par 
Unlike most existing methods, our approach utilizes a part-level semantic layout, decomposing the semantic embedding of each category into global and local components. This decomposition enhances semantic consistency and, when integrated with deep feature modulation, produces higher-quality remote sensing images. Furthermore, few methods, even when employed as enhancements, focus on improving semantic-level diversity and overall performance. To the best of our knowledge, we are the first to introduce the concept of extensibility and employ feature descriptors for remote sensing image synthesis. Our method offers high semantic controllability and diversity, while also exploring enhancements for downstream tasks.
\par 
\section{Method}
Semantic image synthesis aims to generate a photo-realistic image \(I\) with the same height \(H\) and width \(W\) consistent with the given semantic mask \(S \in \mathbb{R}^{H \times W \times C}\), where \(S\) contains \(C\) class labels. Each pixel in \(S\) corresponds to a specific semantic class \(k\) from a set of predefined classes \(\{1, 2, \dots, C\}\), representing the expected semantics at the corresponding location in \(I\).
In the following, we describe the specific methods of HySEGGAN.

\begin{figure*}[htbp]
	\centering
	\includegraphics[scale=0.393]{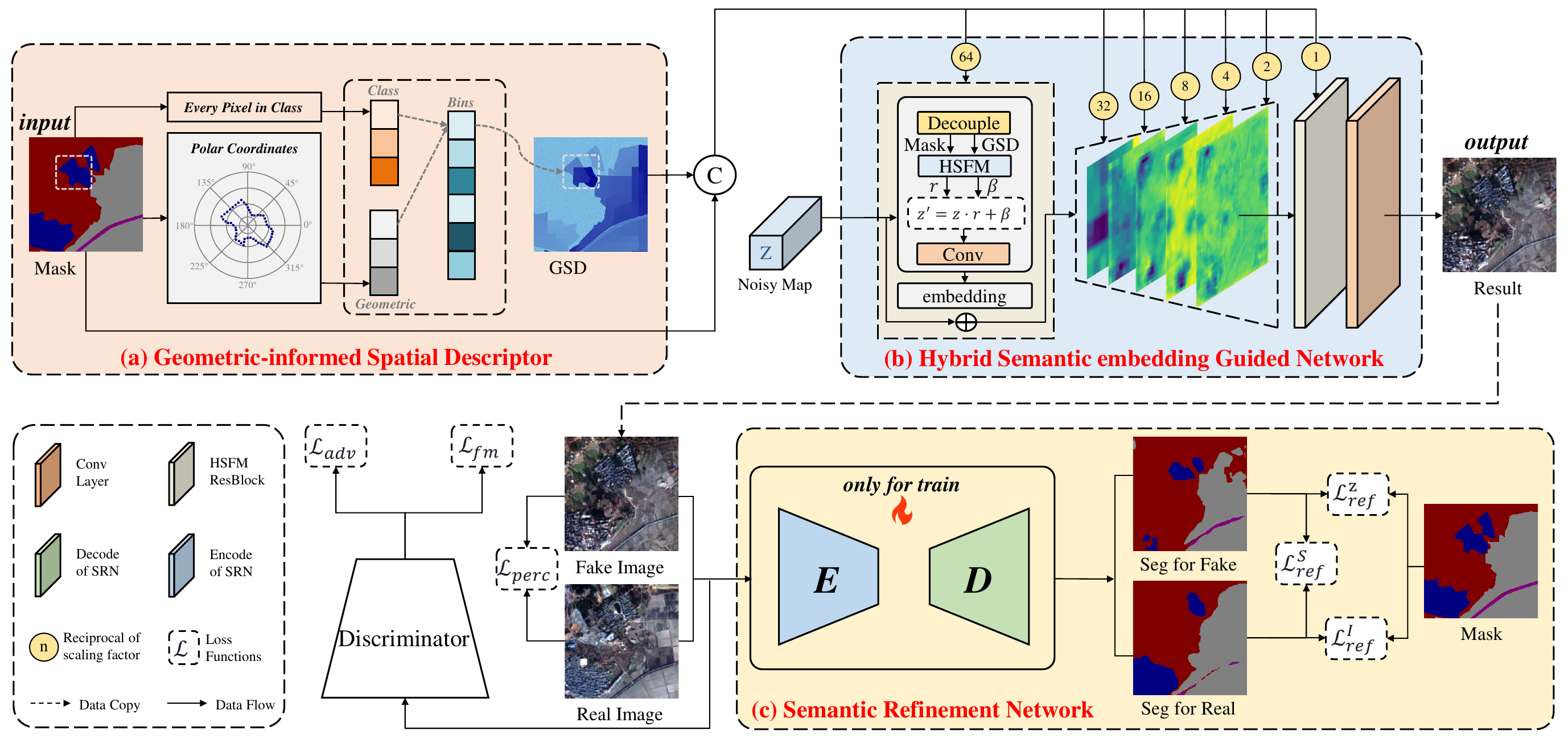}%
	\caption{The training pipeline of the hybrid semantic embedding guided GAN is illustrated as follows. The pink section represents the computation of the Geometric-informed Spatial Descriptor (GSD), the blue section denotes the HSGNet, which primarily includes the HSFMResBlock and upsampling, and the yellow section depicts the semantic refinement network. SRN is based on an encoder-decoder architecture and incorporates a fine-grained semantic feedback process. GSDs are derived from semantic masks, and both are embedded into the input generator using hybrid semantics. The loss functions $\mathcal{L}_{ref}^I$, $\mathcal{L}_{ref}^z$ and $\mathcal{L}_{ref}^S$ proposed for SRN are specifically designed yet trained jointly to enforce semantic consistency constraints and provide fine-grained feedback.}
	\label{fig-pipeline}
\end{figure*}

\subsection{Overall}
\label{Sec-HySEGGAN}
HySEGGAN consists of three main components: GSD, HSGNet and SRN. HSGNet is composed of several HSFMResBlk blocks with upsampling operations. During the generation process, GSDs are calculated from the corresponding semantic embeddings, and multiscale modulation parameters are extracted from the HSFM at each layer. The HSFM then utilizes the modulation parameters to control the generated content via semantic adaptive normalization. The Semantic Refinement Network provides local, fine-grained semantic feedback and semantic mask consistency feedback by using $L_{ref}$, as introduced in Sec.~\ref{Sec-Lref}. The design of the discriminator follows~\cite{tan2022clade} and provides global fidelity feedback.

\subsection{Geometric-informed Spatial Descriptor}
\label{Sec-GSD}
Given \(K\) predefined classes, previous methods typically convert \(S\) to a matrix and apply a uniform convolution kernel over the entire matrix. This simple operation not only limits generative diversity, but also hinders fine-grained controllability and the ability to adapt to complex feature characteristics. Moreover, the method cannot effectively capture the relationships between semantic categories. In addition, this single convolutional operation often leads to low-quality synthetic images that cannot realistically reproduce the shape and texture details of remote sensing objects. This further weakens the model's ability to understand complex scenes.
Informed by the successful application of feature descriptors in natural image processing~\cite{winder2007descriptors,leng2018descriptorsurvy,shakarami2020cbir,lv2022safm}, we propose a geometric-informed spatial descriptor (GSD) that captures positional features of each pixel to describe the geometry and characteristics of remote sensing objects. Additionally, we extend semantic inputs to include one-hot coding and feature descriptors, creating a global-local semantic embedding distribution termed hybrid semantic embedding. This approach allows us to effectively leverage cues from the object part-level layout, thereby enhancing semantic controllability.
\par
\textbf{Calculation of GSD:}
The computational process of GSD can be illustrated using a class in the semantic mask as an example. Based on its segmentation map, the contour shape \( T \in \{0,1\}^{H \times W} \) can be easily obtained, where pixels on the contour are labeled as 1, and the remaining pixels are labeled as 0 by default. Additionally, the contour shape of a class can be discretely represented as the set of points \( P = \{(x_i, y_i) \mid T(x_i, y_i) = 1\} \). For any point \( S = (x_s, y_s) \) within an instance, we compute its GSD features through the following steps.
For a point \( S = (x_s, y_s) \), treat it as the pole and construct a polar coordinate space around it. Describe the position of a point \( p = (x_i, y_i) \) on the contour line relative to \( S \) in polar coordinates. For each point \( p \) on the contour, calculate its Euclidean distance $r_i$ and polar angle $\theta_i$ to \( S \), as given by Eq.~\eqref{distance of points} and Eq.~\eqref{polar of angle}.
\begin{equation}
\label{distance of points}
r_{i} = \sqrt{\left. \left( x \right._{i} - x_{s} \right)^{2} + \left. \left( y \right._{i} - y_{s} \right)^{2}}
\end{equation}
\begin{equation}
\label{polar of angle}
\theta_{i} = arctan\left( -(y_{i} - y_{s}),~x_{i} - x_{s} \right)
\end{equation}
Note that we take the negative of $\Delta y$ because, in the image coordinate system, the $y$-axis increases downward, and we want the angle calculation to follow the traditional polar coordinate definition. Here, $arctan$ represents the arctangent function, and the angle ranges from $-\pi$ to $\pi$. To ensure that the angle $\theta_i$ lies within the range $[0,2\pi)$, we adjust it as shown in Eq.~\eqref{normalizing of the angle}.
\begin{equation}
\label{normalizing of the angle}
{\theta_{i}}^{'} = \theta_{i} + 2\pi\left( \theta_{i} < 0 \right)
\end{equation}
Calculate the maximum distance among all, and then normalize each distance to $r'$, as given by Eq.~\eqref{normalizing of the distance}.
\begin{equation}
\label{normalizing of the distance}
{r_{i}}^{'} = \frac{2r_{i}}{R_{max}}
\end{equation}
Let $k \in \{{1, 2, \dots, n_\rho}\}$. We create $n_\rho$ radial boundaries, ${r_k}$, using logarithmic spacing, as shown in Eq.~\eqref{radial b}.
\begin{equation}
\label{radial b}
r_k = \text{logspace} \left( \log_{10} r_{\text{inner}},\ \log_{10} r_{\text{outer}},\ n_\rho\right)_k
\end{equation}
Where $r_{\text{inner}}$ and $r_{\text{outer}}$ represent the preset values for the inner and outer radii, respectively. The angles are uniformly divided into $n_\theta$ intervals, each with a size given by Eq.~\eqref{intervals}.
\begin{equation}
\label{intervals}
\Delta \theta = \frac{2\pi}{n_\theta}
\end{equation}
For each normalized distance $\tilde{r}_i$, determine its corresponding radial bin, as given by Eq.~\eqref{radial bin}.
\begin{equation}
\label{radial bin}
b_{r,i} = \min \left\{ k \mid\ r_{i}^{'} \leq r_k \right\}
\end{equation}
For each adjusted angle $\theta_i'$, compute its corresponding angular bin, as given by Eq.~\eqref{angle bin}.

\begin{equation}
\label{angle bin}
b_{\theta,i} = 1 + \left \lfloor \frac{\theta_i'}{\Delta \theta} \right\rfloor,\quad \theta_i' \in [0,\ 2\pi)
\end{equation}
Where $\left\lfloor \cdot \right\rfloor$ denotes the floor operation.
Initialize a shape descriptor histogram $H \in \mathbb{R}^{n_r \times n_\theta}$, with all elements initialized to zero.
For each contour point $p_i$, update the histogram as shown in Eq.~\eqref{count}.
\begin{equation}
\label{count}
H(b_{r,i},\ b_{\theta,i}) = H(b_{r,i},\ b_{\theta,i}) + 1
\end{equation}
Compute the total sum $N$ of the histogram elements and normalized $H$ as shown in Eq.~\eqref{normalize}.
\begin{equation}
\label{normalize}
N = \sum_{k=1}^{n_r} \sum_{\ell=1}^{n_\theta} H(k,\ \ell), \quad H_{\text{norm}} = \frac{H}{N + \epsilon}
\end{equation}
Where $\epsilon$ is a very small value used to avoid division by zero, $\epsilon = 1e-8$ in this paper. For each instance in the image, the above process is repeated to compute the GSD features for all pixels within the instance.
Compute the global mean $\mu$ and standard deviation $\sigma$ of all descriptors, as shown in Eq.~\eqref{mu} and Eq.~\eqref{sigma}, and then standardize the descriptors, as shown in Eq.~\eqref{std}.
\begin{equation}
\label{mu}
\mu = \frac{1}{M} \sum_{i=1}^{M} H_{\text{norm},\ i}
\end{equation}
\begin{equation}
\label{sigma}
\sigma = \sqrt{ \frac{1}{M} \sum_{i=1}^{M} \left( H_{\text{norm},\ i} - \mu \right)^2 }
\end{equation}
\begin{equation}
\label{std}
H_{\text{std},\ i} = \frac{H_{\text{norm},\ i} - \mu}{\sigma + \epsilon}
\end{equation}
Where $M$ denotes the total number of descriptors. The pseudocode for the geometric-informed spatial descriptor is shown in Algorithm~\ref{alg-GSD}. To balance the computational cost and robustness of the descriptors, we use only the contour points of an object instance to describe its shape information, rather than using all the points within the segmentation region.

\begin{algorithm}[!t]
	\caption{Pseudocode of geometric-informed spatial descriptor} 
	\label{alg-GSD}
	\begin{algorithmic}[1]
		\Require Input instance $I$ of shape $(bsz, _, h, w)$
		\Ensure
		Descriptor tensor $D$ of shape $(bsz, n_{\theta} \times n_{\rho}, h, w)$
    	\State Initialize descriptor tensor $D \leftarrow \text{zeros}(bsz, n_{\theta} \times n_{\rho}, h, w)$
     
    	\For{each batch $b\_num \in [0, bsz)$}
    		\State Extract instance indices $U \leftarrow \text{unique}(I[b\_num])$
    		
    		\For{each index $idx \in U$} 
    			\If{$idx \neq 0$} 
    				\State Create binary mask $M \leftarrow (I[b\_num] == idx)$
    				\State Find contours from mask $M \rightarrow C$ 
    				\State Create binary image $R \leftarrow \text{drawContours}(C)$
    				\State Get coordinates of contour pixels $C\_p$ and highlight pixels $H\_p$
    
    				\State $S \leftarrow \text{concat}(C\_p, H\_p)$ \
    				
    				\State $\text{Compute}(C\_p, S, D[b\_num])$ \
    			\EndIf
    		\EndFor	
    	\EndFor
     
        \State $mean \leftarrow \text{mean}(D, \text{dim}=[1, 2, 3], \text{keepdim}=\text{True})$ \
        \State $std \leftarrow \text{std}(D, \text{dim}=[1, 2, 3], \text{keepdim}=\text{True})$ \
        \State $N \leftarrow std + \epsilon$ \
        \State $normalized\_D \leftarrow \frac{D - mean}{N}$ \
        
    	\State \textbf{Return} $normalized\_D$ \

    \end{algorithmic}  
\end{algorithm}

\subsection{Hybrid Semantic Embedding Guided Network}
\par
\begin{figure}[!t]
\centering
\includegraphics[scale=0.45]{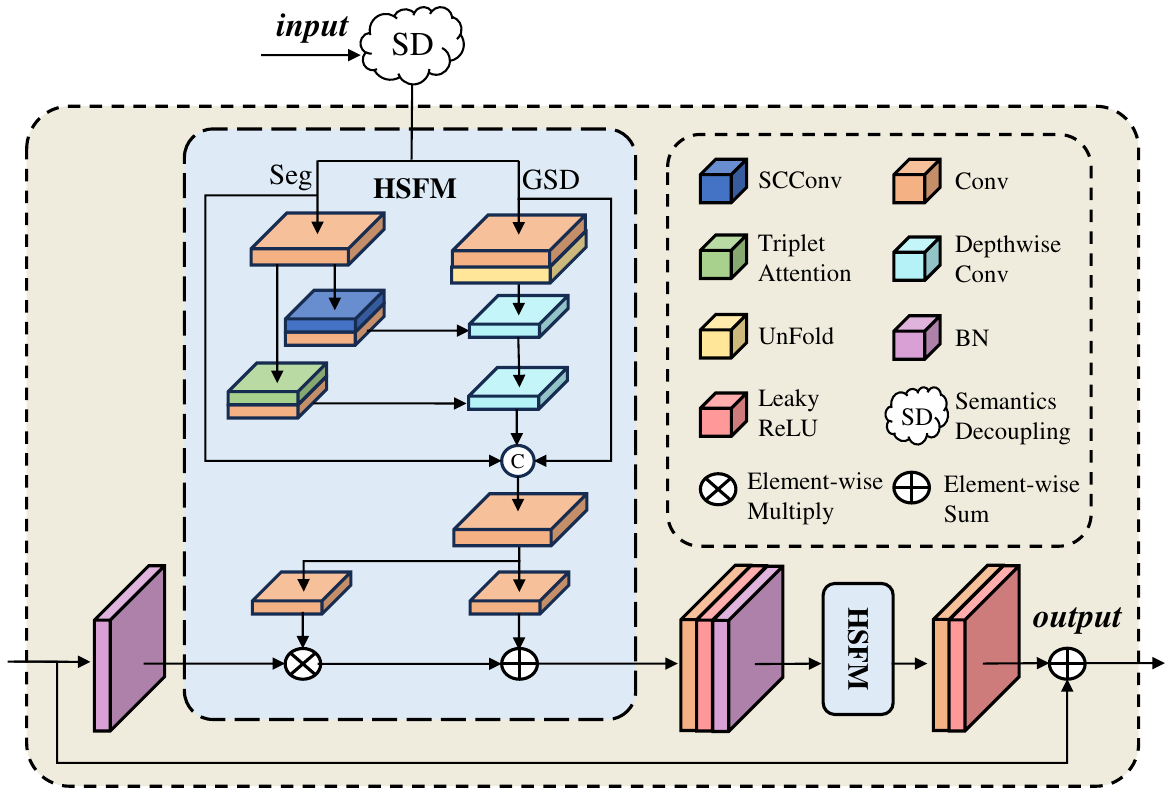}
\caption{Detailed structure of the HSFMResBlock used in Fig.~\ref{fig-pipeline}. It learns pixel-level fine-grained modulation parameters from the hybrid semantic embedding and guides the modulation of normalized activations.}
\label{fig-hsfm}
\end{figure}

We call this network Hybrid Semantic embedding Guided Network (HSGNet). The architecture of HSGNet is shown in Fig.~\ref{fig-pipeline}. It is observed that the complex spatial characteristics of remote sensing objects are predominantly reflected in the variability of textures within the same category and the similarity of shapes across different categories. To mitigate this issue, we design the Hybrid Semantic Feature Modulation (HSFM) block in Fig.~\ref{fig-hsfm}, which combines semantic information and our GSD features to compensate each other, inspired by~\cite{ResNet,chollet2017deepconv,misra2021triplet,li2023scconv,lv2022safm}. 
\par
Specifically, the GSD features supplement the object-level geometric shape and spatial information for the semantic layout, while the semantic layout introduces pairwise global semantic information to the GSD features. HSFM then adaptively adjusts the parameters based on the varying shapes of different semantic object classes, effectively guiding the semantic image synthesis.
In the HSFM block, the hybrid semantic features are initially decoupled, and the resulting semantic layout and GSD features are scaled to a uniform size. These features are then sent to two separate convolutional layers to predict two sets of semantically adaptive 3 × 3 convolutional kernels that vary according to the class label of each spatial location. The features obtained after the convolution of the semantic layout are then passed through TripletAttention~\cite{misra2021triplet} and SCConv~\cite{li2023scconv}, respectively, to reduce spatial and channel redundancy within the features of the convolutional neural network, thereby compressing the CNN model and enhancing its performance. Afterward, two sets of features are obtained.
The semantic information of each spatial location is then gradually fused with the corresponding location in the GSD features through deep convolutional layers.
\par
Finally, feature modulation is carried out using hybrid semantic features and adaptive normalized modulation parameters. By using HSFM blocks, semantic and spatial location information can be effectively integrated. The extensive prior knowledge embedded in the hybrid semantic embedding significantly enhances both semantic consistency and the fine-grained generation quality of images synthesized by HSGNet, especially for remote sensing targets, resulting in exceptional performance.

\subsection{Semantic Refinement Network}
\label{Sec-HSFM}
In the traditional generative adversarial network (GAN) training framework, the generator and discriminator networks compete with each other. The generator produces images, while the discriminator seeks to differentiate between real and fake images. Recent research has focused on designing new discriminator architectures~\cite{sushko2020discriminator} that better incorporate semantic masks. However, this requires the discriminator to simultaneously evaluate both image fidelity and semantic mask consistency, significantly increasing its learning complexity and potentially hindering the generator from receiving fine-grained semantic feedback.
\par
\begin{figure}[!t]
\centering
\includegraphics[scale=0.77]{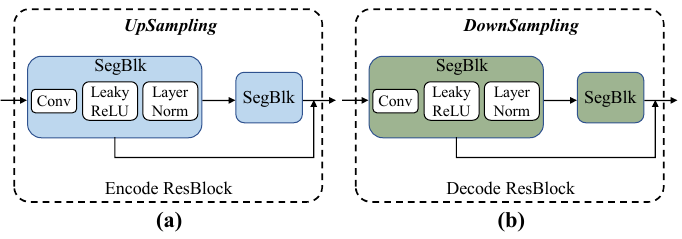}
\caption{Illustrate the architecture of encoder layers and decoder layers in semantic refinement networks as shown in Fig.~\ref{fig-SRN}(a) and Fig.~\ref{fig-SRN}(b), respectively.}
\label{fig-SRN}
\end{figure}

To address this challenge, we introduce a novel semantic refinement network (SRN) into the traditional training framework. During training, the discriminator provides global fidelity feedback, while the SRN offers local, fine-grained semantic and mask consistency feedback (Fig. ~\ref{fig-SRN}). The SRN is designed as a semantic segmentation network with an encoder-decoder architecture that classifies each pixel of the input image. The pixel-level cross-entropy loss is computed between the SRN's output and the ground truth segmentation masks. The SRN is jointly trained with the generator and discriminator. When provided with real images, the SRN performs self-training and generates alignment information. In contrast, for fake images, the SRN delivers segmentation results and alignment data as fine-grained semantic feedback. The losses $L_{ref}^z$ and $L_{ref}^S$ are activated after the warm-up segmentation epoch, as described in Sec.~\ref{Sec-Lref}.

\subsection{Loss Function}
In this approach, we utilize adversarial loss $L_{adv}$, feature matching loss $L_{fm}$ and perceptual loss $L_{perc}$ to enhance both fidelity and controllability of the generated images. Additionally, we introduce $L_{ref}$ to further improve the controllability of the generator.
\subsubsection{Adversarial loss}
\par
Adversarial learning is effective in maintaining the synthesized image's alignment with the real image distribution and is widely used in various image generation tasks~\cite{brock2018large,mirza2014conditional}. In this work, we employ a hinge-based adversarial loss~\cite{zhang2019selfattn,miyato2018spectralnorm}, and the optimization process for the generator $G$ and discriminator $D$ is formulated as shown in Eq.~\eqref{L_adv_d} and Eq.~\eqref{L_adv_g}.
\begin{equation}
\begin{aligned}
\label{L_adv_d}
\mathcal{L}_{adv}^D=-\mathbb{E}_{\left(I,S\right)}\left[\min{\left(0,-1+D\left(I,S\right)\right)}\right]-\\\mathbb{E}_{\left(z,S\right)}\left[min\left(0,-1-D\left(G\left(z,S\right),S\right)\right)\right]\
\end{aligned}
\end{equation}
\begin{equation}
\label{L_adv_g}
\mathcal{L}_{adv}^G=-\mathbb{E}_{\left(z,S\right)}[D\left(G\left(z,S\right),S\right)]
\end{equation}
Where $I$ and $S$ represent the real image and the semantic label of the input, respectively, and $z$ is the hidden variable of the input generator.
\subsubsection{Feature Matching loss}
Following~\cite{wang2018p2phd}, we employ the feature matching loss $L_{fm}$ to enhance supervision and stabilize the training process. This loss encourages the features of the synthesized image to be closer to those of the real image in the feature space of discriminator $D$. The feature matching loss $L_{fm}$ is defined as shown in Eq.~\eqref{L_fm}.
\begin{equation}
\label{L_fm}
\mathcal{L}_{fm}=\sum_{i=1}^{n}\frac{1}{N_i}[\vert\vert\left({D}_i\left(I,S\right)-D_i\left(G\left(z,S\right),S\right)\right\vert\vert_1]
\end{equation}
Where $ N_i$ is the number of elements in the feature $D_i(I, S)$.
\subsubsection{Perceptual loss} 
We use a pre-trained VGG-19 model $\Phi$~\cite{simonyan2014vgg} to extract features from the real image $I$ and synthesized image respectively. The perceptual loss is computed in the multi-scale feature space as shown in Eq.~\eqref{L_perc}.
\begin{equation}
\label{L_perc}
\mathcal{L}_{perc}=\sum_{m=1}^{M}[\vert\vert\Phi_k(\hat{I})-\Phi_k\left(I\right)\vert\vert_1]
\end{equation}
The $m$-th feature map, denoted as $\Phi_k$, indicates the extraction from the VGG-19 model, where the subscript $k$ represents the $k$-th layer of the network. We set $M=5$ in this paper.
\subsubsection{Semantic Refinement loss}
\label{Sec-Lref}
To ensure a high degree of semantic consistency between the synthesized image and the given semantic mask, we propose the semantic refinement loss to optimize the training process of model, as shown in Eq.~\eqref{L_ref}.
\begin{equation}
\label{L_ref}
\mathcal{L}_{ref}(e)=\begin{cases}
\mathcal{L}_{ref}^I, & 0\leq e < \Gamma
\\\\
\mathcal{L}_{ref}^I+\mathcal{L}_{ref}^z+\mathcal{L}_{ref}^S, & e \geq \Gamma
\end{cases}
\end{equation}
Where, $e$ represents the current training epoch, and $\Gamma$ is a hyperparameter. Additionally, $\mathcal{L}_{ref}^I$ represents the loss function for real images, $\mathcal{L}_{ref}^z$ corresponds to the loss function for synthesized images, and $\mathcal{L}_{ref}^S$ is the loss function for both real and synthesized images. The definition of $\mathcal{L}_{ref}^I$ is given in Eq.~\eqref{L_ref_I}.
\begin{equation}
\label{L_ref_I}
\mathcal{L}_{ref}^I = - \mathbb{E} \left[ \sum_{j=1}^{M} \sum_{k=1}^{K} m_{j,k} \log\left( \frac{\exp(C_k(I_j))}{\sum_{i=1}^{K} \exp(C_i(I_j))} \right) \right]
\end{equation}
Where \(K\), \(I\) and \(C_k \) represent the total number of semantic categories, the real image and the output of the semantic refinement network for the \(k-th\) class, respectively, and \(m\) is the input semantic mask. The definition of ${L}_{ref}^z$ is shown in Eq.~\eqref{L_ref_z}.
\begin{equation}
\label{L_ref_z}
\mathcal{L}_{ref}^z = -\mathbb{E}\left[ \sum_{j=1}^{M} \sum_{k=1}^{K} m_{j,k} \log \left( \frac{\exp(C_k(G(z)_j))}{\sum_{i=1}^{K} \exp(C_i(G(z)_j))} \right) \right]
\end{equation}
Where $G(z)$ denotes the image synthesized by the generator, as defined in Eq.~\eqref{L_ref_S}.
\begin{equation}
\label{L_ref_S}
\mathcal{L}_{ref}^S = - \mathbb{E} \left[ \sum_{j=1}^{M} \sum_{k=1}^{K} Pre(I_{j,k}) \log \left( Pre(z_{j,k}) \right) \right]
\end{equation}
Where $Pre()$ represents the result of the semantic refinement network.
\par
Despite $\mathcal{L}_{ref}^I$, $\mathcal{L}_{ref}^z$ and $\mathcal{L}_{ref}^S$ share formal similarities, they each produce distinct effects. $\mathcal{L}_{ref}^I$ aligns the segmentation results of real images with the given mask, facilitating the training of the semantic refinement network. $\mathcal{L}_{ref}^z$ leverages the segmentation results generated by the semantic refinement network for synthesized images, feeding them back to the generator via backward propagation to guide image synthesis. $\mathcal{L}_{ref}^S$ further refine the generator's output by minimizing the discrepancy between the segmentation results of synthesized images and real images, thereby improving the semantic consistency and quality of the synthesized images.
\subsubsection{Total loss}
The generator is trained using the adversarial loss $\lambda_{adv}$, the feature matching loss $\lambda_{fm}$, the perceptual loss $\lambda_{perc}$ and the proposed semantic refinement loss $\lambda_{ref}$. The total loss is shaped as shown in Eq.~\eqref{L_total}.

\begin{equation}
\begin{aligned}
\label{L_total}
\mathcal{L} = \min_G \Big( \max_D \left( \mathcal{L}_{adv}^D \right) + \lambda_{adv} \mathcal{L}_{adv}^G + 
\\ \lambda_{fm} \mathcal{L}_{fm} + \lambda_{perc} \mathcal{L}_{perc} + \lambda_{ref} \mathcal{L}_{ref} \Big)
\end{aligned}
\end{equation}
Where $\lambda_{adv}$, $\lambda_{fm}$, $\lambda_{perc}$ and $\lambda_{ref}$ are trade-off parameters.

\section{Experiments}
We compare our approach with several state-of-the-art semantic image synthesis methods, including Pix2PixHD~\cite{wang2018p2phd}, SPADE~\cite{park2019spade}, CLADE~\cite{tan2022clade}, SAFM~\cite{lv2022safm}, INADE~\cite{tan2023inade} and Santa~\cite{xie2023santa}. Among these, Pix2PixHD, SPADE and SAFM are quality-oriented, while CLADE and INADE emphasize diversity. Santa, on the other hand, represents methods based on unmatched data and is thus evaluated using only FID and its improvement on downstream tasks. All models were trained on remote sensing datasets using open-source code provided by the authors. Bold and underlined fonts indicate the optimal and sub-optimal values for each metric. Methods that do not support diversity synthesis are excluded from the evaluation, and their corresponding values are replaced by "-".
\subsection{Experimental Settings}
\subsubsection{Dataset}
The experiments were conducted across various remote sensing scenarios, and the following datasets were used: 
(1) GID~\cite{GID2020}, a large-scale land cover dataset consisting of images acquired by the Gaofen-2 (GF-2) satellite. GID-15 contains 16 semantic categories, including $industrial$ $land,$ $urban$ $residential,$ $rural$ $residential,$ $traffic$ $land,$ $paddy$ $field,$ $irrigated$ $land,$ $dry$ $cropland,$ $garden$ $plot,$ $arbor$ $woodland,$ $shrub$ $land,$ $natural$ $grassland,$ $artificial$ $grassland,$ $river,$ $lake,$ $pond$ and $background.$ We used the fine land cover classification component, GID-15, which consists of 20 remote sensing images (6800 × 7200) along with their corresponding semantic masks. These images were divided into 256 × 256 image blocks, without overlap or intersection. The selected 2114 paired images were randomly split into training and validation sets at a 6:1 ratio, resulting in 1812 training images and 302 validation images.
(2) ISPRS~\cite{rottensteiner2014isprs} is the classic aerial imagery dataset, of which Potsdam is used in this paper. Potsdam contains 38 images of 6000 × 6000 size. The semantic categories in them are categorized into six land cover categories, i.e., $impervious$ $surface,$ $buildings,$ $low$ $vegetation,$ $trees,$ $cars$ and $background.$ For training purposes, we split each image into 256 × 256 images sequentially.
Details of these datasets are as shown in Table~\ref{table-dataset}.
\begin{table}[!t]
    \caption{Details of the dataset used for the experiment.\label{table-dataset}}
    \centering
    \begin{tabular}{c|ccccc}
        \toprule
        Dataset  & Resolution & Type & Class & Train-num & Val-num \\
        \midrule
        GID-15 & 0.8m & Satellite & 16 & 1812 & 302 \\
        ISPRS & 0.05m & Airborne & 6 & 2712 & 462 \\
        \bottomrule
    \end{tabular}
\end{table}
\subsubsection{Experimental Details}
Our model is implemented using PyTorch 1.5.0. The experimental platform is a Linux system, and the experiments are executed on a single GeForce RTX 4090 GPU. We use the Adam~\cite{kingma2014adam} optimizer, setting $\beta_1 = 0$ and $\beta_2 = 0.999$, and train using the TTUR~\cite{heusel2017ttur} strategy. The initial learning rate is set to 0.0001 for the generator, 0.0004 for the discriminator, and 0.0001 for the semantic refinement network. The parameters $\lambda_{adv}$, $\lambda_{fm}$, $\lambda_{perc}$ and $\lambda_{ref}$ are set to 1, 10, 10 and 1, respectively. The hyperparameters are set as $n_\rho = 6$, $n_\theta = 12$, and $\Gamma = 80$, according to the ablation study in Sec.~\ref{Sec-ablation-hyper}. All models are trained for 200 epochs, during which the learning rate decays to zero in the last 100 epochs. The training environment and parameter settings are kept the same for all models, and there are no outputs requiring any specific post-processing operations.
\subsubsection{Controllbility Metrics}
Following previous semantic image synthesis work~\cite{park2019spade,lv2022safm}, the Inception-V3~\cite{xia2017inception} network is used in the FID~\cite{bynagari2019fid} calculation to extract features from real and synthesized images. The last fully connected or pooled layer is removed to obtain a 2048-dimensional feature vector. After this, the mean and covariance matrices are used to calculate the distance between the two high-dimensional distributions to compute the FID score as shown in Eq.~\eqref{FID}.
\begin{equation}
\label{FID}
FID\left(x,g\right)=||\mu_x-\mu_g||_2^2+T_r(C_x+C_g+2(C_xC_g))^{1/2}
\end{equation}
Where $\mu_x$ and $\mu_g$ represent the mean values of the real and synthesized images, respectively, while $C_x$ and $C_g$ denote the covariance matrices of the real and synthesized images, respectively. 
For segmentation-based metrics, we evaluated the mean intersection over union and pixel accuracy of pre-trained semantic segmentation models on synthesized images. Higher fidelity leads to better performance in a well-trained segmentation model. Let $n$ represent the total number of categories, and let $p_{ij}$ denote the probability that category $i$ is predicted as $j$. The formulas for the mean intersection over union and pixel accuracy are presented in Eq.~\eqref{mIoU} and Eq.~\eqref{acc}.
\begin{equation}
\label{mIoU}
mIoU=\frac{1}{N}\sum_{i=0}^{N}\frac{p_{ii}}{\sum_{j=0}^{j=N}p_{ji}+\sum_{j=0}^{j=N}{p_{ij}-p_{ii}}}\ 
\end{equation}
\begin{equation}
\label{acc}
Accuracy=\frac{\sum_{i=0}^{N}p_{ii}}{\sum_{i=0}^{N}\sum_{j=0}^{j=N}p_{ij}}
\end{equation}
We used DeeplabV3+~\cite{DeepLabv3+} as the segmentation model and trained it for 200 epochs.
\subsubsection{Diversity Metrics}
We used LPIPS~\cite{zhang2018lpips} to assess global diversity. LPIPS measures the weighted L1 distance. In order to evaluate diversity multi-dimensionally, we synthesize 5 images separately for each semantic mask and randomly sample 5 pairs of images for each mask to calculate the average LPIPS as the final result. Higher LPIPS scores indicate that the synthesized images have better global diversity. Following~\cite{tan2023inade,wang2023csebgan} in addition, we use mean Class-Specific Diversity (mCSD) and mean Other-Classes Diversity (mOCD) proposed by~\cite{zhu2020groupd}. Given a set of images $S={\{I_{1}^1, ..., I_{1}^n..., I_{C}^1, ..., I_{C}^n}\}$. mCSD and mOCD are calculated by Eq.~\eqref{mcsd} and Eq.~\eqref{mocd}.

\begin{equation}
\label{mcsd}
mCSD =\frac{1}{C}\sum_{c=1}^{C}{L_{c}}\ 
\end{equation}

\begin{equation}
\label{mocd}
mOCD =\frac{1}{C}\sum_{c=1}^{C}{L_{\neq c}}\ 
\end{equation}
Where $L_{c}$ is the average LPIPS distance~\cite{zhang2018lpips} of the semantic area of class $c$ between sampled $m$ pairs while $L_{\neq c}$ represents the average LPIPS distance in the areas of all other classes between the same pairs. 
mCSD quantifies the diversity of specific semantic categories, while mOCD evaluates the stability of other categories. Thus, effective diversity necessitates high diversity in specific semantic domains (high mCSD) and low diversity in all other domains (low mOCD).

\subsubsection{Extensibility Metrics}
To evaluate the performance improvement brought by the synthetic images for downstream segmentation tasks, we first train a baseline model on the original training set. And then compare its performance with that of the model trained on an augmented dataset, which combines the original training set with synthetic images. We use DeeplabV3+ as the architecture, trained with the Adam optimizer for the same number of iterations in both augmented and unaugmented settings. The performance metrics include mIoU, FWIoU and accuracy, which we denote as d-mIoU, d-FWIoU and d-acc for the downstream remote sensing image segmentation tasks. The formulas for d-mIoU and d-acc are provided in Eq.~\eqref{mIoU} and Eq.~\eqref{acc}, respectively, while the formula for d-FWIoU is given in Eq.~\eqref{FWIoU}.
\begin{equation}
\label{FWIoU}
FWIoU=\sum_{i=0}^{N}{w_i}\cdot \frac{p_{ii}}{\sum_{j=0}^{j=N}p_{ji}+\sum_{j=0}^{j=N}{p_{ij}-p_{ii}}}\ 
\end{equation}
The weight \( w_i \) for class \( i \) is calculated as the ratio of the number of pixels in the true label for that class to the total number of pixels.
\subsection{Quantitative Comparison}
\begin{table*}[!t]
    \begin{minipage}{0.48\textwidth}
        \caption{Quantitative comparison on GID-15 task. The optimal and suboptimal results are highlighted in bold and underlined, respectively.\label{table-comparison-1}}
        \centering
        \setlength{\tabcolsep}{0.85mm}{
        \begin{tabular}{l|ccc|ccc}
            \toprule
            \multirow{2}*{Method} &
            \multicolumn{3}{c|}{Controllability} & \multicolumn{3}{c}{Diversity} \\
            & FID$\downarrow$ & mIOU$\uparrow$ & acc$\uparrow$ & LPIPS$\uparrow$ & mCSD$\uparrow$ & mOCD$\downarrow$ \\
            \midrule
            Pix2PixHD~\cite{wang2018p2phd} & 206.114 & 49.930 & 74.217 & 0 & - & - \\
            SPADE~\cite{park2019spade} & 142.579 & 57.044 & 82.673 & 0 & - & - \\
            SAFM~\cite{lv2022safm} & 139.791 & \uline{64.318} & \uline{86.601} & 0.151 & 0.019 & \uline{0.082}  \\
            CLADE~\cite{tan2022clade} & 145.655 & 59.591 & 81.069 & 0.187 & 0.027 & 0.105 \\
            INADE~\cite{tan2023inade} & \uline{136.615} & 52.074 & 78.774 & \textbf{0.339} & \textbf{0.054} & 0.200  \\
            Santa~\cite{xie2023santa} & 157.844 & - & - & 0 & - & - \\
            \midrule
            \textbf{Ours} & \textbf{132.283} & \textbf{71.234} & \textbf{88.022} & \uline{0.189} & \uline{0.028} & \textbf{0.073} \\
            \bottomrule
        \end{tabular}}
    \end{minipage}
    \hfill
    \begin{minipage}{0.48\textwidth}
        \caption{Quantitative comparison on ISPRS task. The optimal and suboptimal results are highlighted in bold and underlined, respectively.\label{table-comparison-2}}
        \centering
        \setlength{\tabcolsep}{0.85mm}{
        \begin{tabular}{l|ccc|ccc}
            \toprule
            \multirow{2}*{Method} &
            \multicolumn{3}{c|}{Controllability} & \multicolumn{3}{c}{Diversity}  \\
            & FID$\downarrow$ & mIOU$\uparrow$ & acc$\uparrow$ & LPIPS$\uparrow$ & mCSD$\uparrow$ & mOCD$\downarrow$ \\
            \midrule
            Pix2PixHD~\cite{wang2018p2phd} & 185.589 & 58.571 & 80.010 & 0 & - & -  \\
            SPADE~\cite{park2019spade} & \uline{144.340} & 67.168 & 86.382 & 0 & - & -  \\
            SAFM~\cite{lv2022safm} & 156.077 & 65.902 & 86.510 & \uline{0.349} & 0.041 & \uline{0.122} \\
            CLADE~\cite{tan2022clade} & 153.128 & \uline{68.413} & \uline{87.537} & 0.335 & 0.043 & 0.123  \\
            INADE~\cite{tan2023inade} & 146.597 & 62.868 & 83.094 & \textbf{0.419} & \textbf{0.070} & 0.197  \\
            Santa~\cite{xie2023santa} & 173.354 & - & - & 0 & - & - \\
            \midrule
            \textbf{Ours} & \textbf{134.195} & \textbf{77.403} & \textbf{91.98} & 0.317 & \uline{0.044} & \textbf{0.099}  \\
            \bottomrule
        \end{tabular}}
    \end{minipage}
\end{table*}

\begin{table*}[h]
    \caption{Quantitative comparison on extensibility, results from GID-15 and ISPRS tasks. Source only indicates no synthesized images used, and Baseline is the benchmark for removing the three modules. The optimal and suboptimal results are highlighted in bold and underlined, respectively.\label{table-comparison3}}
    \centering
    \setlength{\tabcolsep}{3.5mm}{
    \begin{tabular}{l|ccc|ccc}
        \toprule
        \multirow{2}*{Method} &
		\multicolumn{3}{c|}{GID-15} & \multicolumn{3}{c}{ISPRS} \\
        & d-mIoU↑ & d-FWIoU↑ & d-acc↑ & d-mIoU↑ & d-FWIoU↑ & d-acc↑\\
        \midrule
        Source only & 56.867 (+0.000) & 68.250 (+0.000) & 80.814 (+0.000) & 67.102 (+0.000) & 79.148 (+0.000) & 87.107 (+0.000) \\
        \midrule
        Pix2PixHD~\cite{wang2018p2phd} & 55.236 (-1.631) & 68.644 (+0.394) & 81.151 (+0.337) & 66.950 (-0.152) & 76.968 (-2.180) & 87.350 (+0.243) \\
        SPADE~\cite{park2019spade} & 58.341 (+1.474) & 69.421 (+1.171) & 81.682 (+0.868) & 67.960 (+0.858) & 79.450 (+0.302) & 87.848 (+0.741) \\
        SAFM~\cite{lv2022safm} & 56.179 (-0.688) & 68.903 (+0.653) & 81.365 (+0.551) & 67.113 (+0.011) & 79.931 (+0.783) & 87.670 (+0.563) \\
        CLADE~\cite{tan2022clade} & \uline{58.353 (+1.486)} & 69.522 (+1.272) & 81.724 (+0.910) & \uline{68.250 (+1.148)} & 80.222 (+1.074) & \uline{88.020 (+0.913)} \\ 
        INADE~\cite{tan2023inade} & 57.681 (+0.814) & \uline{69.713 (+1.463)} & \uline{81.873 (+1.059)} & 67.903 (+0.801) & \uline{80.480 (+1.332)} & 87.950 (+0.843) \\
        Santa~\cite{xie2023santa} & 55.202 (-1.665) & 68.263 (+0.013) & 80.834 (+0.020) & 68.105 (+1.003) & 80.123 (+0.975) & 88.009 (+0.902) \\
        \midrule
        \textbf{Ours} & \textbf{59.191 (+2.324)} & \textbf{69.747 (+1.497)} & \textbf{82.002 (+1.188)} & \textbf{68.480 (+1.378)} & \textbf{80.389 (+1.241)} & \textbf{88.240 (+1.133)} \\
        \bottomrule
    \end{tabular}}
\end{table*}

\subsubsection{Synthesis results on GID-15} Table~\ref{table-comparison-1} compares the performance of our method with competing approaches on GID-15~\cite{GID2020} using rigorous experimental protocols. In the semantic controllability experiments, we compare several state-of-the-art methods across two dimensions: generation quality (primarily measured by FID~\cite{bynagari2019fid}) and semantic consistency (assessed by mIoU and accuracy). Compared to Pix2PixHD~\cite{wang2018p2phd}, SPADE~\cite{park2019spade}, SAFM~\cite{lv2022safm}, CLADE~\cite{tan2022clade}, INADE~\cite{tan2023inade} and Santa~\cite{xie2023santa}, our method significantly outperforms the others in terms of FID, mIoU and accuracy. This demonstrates that our method has significant advantages in both the quality and semantic consistency of synthesized images. Diversity experiments show that our method achieves sub-optimal performance on LPIPS and mCSD metrics, and optimal performance on the mOCD metric. This demonstrates that our method synthesized high-quality, semantically consistent images while also excelling in maintaining generative diversity, as confirmed by the diversity metrics.
\subsubsection{Synthesis results on ISPRS}
Table~\ref{table-comparison-2} presents a comparison of our method with other approaches. On the ISPRS~\cite{rottensteiner2014isprs} dataset, our method achieves state-of-the-art performance in controllability, significantly outperforming other methods. This highlights the robustness of our approach across different remote sensing scenarios. Regarding diversity, the lower spatial resolution and limited semantic categories of the ISPRS dataset somewhat constrain the diversity of synthesized images in terms of semantic features, leading to poor LPIPS performance. Nevertheless, mCSD and mOCD achieve sub-optimal and optimal results, respectively.
\subsubsection{Downstream tasks on both GID-15 and ISPRS}
Table~\ref{table-comparison3} presents a comparison of different methods in terms of extensibility. Our method performs well across all three metrics. The results demonstrate that our method excels in accuracy, extensibility and stability, effectively supporting downstream tasks in remote sensing while ensuring both semantic controllability and diversity. The significant advantages of our method are validated through experiments on GID-15 and ISPRS, two remote sensing image datasets with vastly different characteristics. This indicates that our method is both robust and adaptable, making it suitable for data enhancement applications across a variety of remote sensing image processing tasks.
\subsection{Qualitative Comparison}
\begin{figure*}[htbp]
\centering
{\includegraphics[width=7.18in]{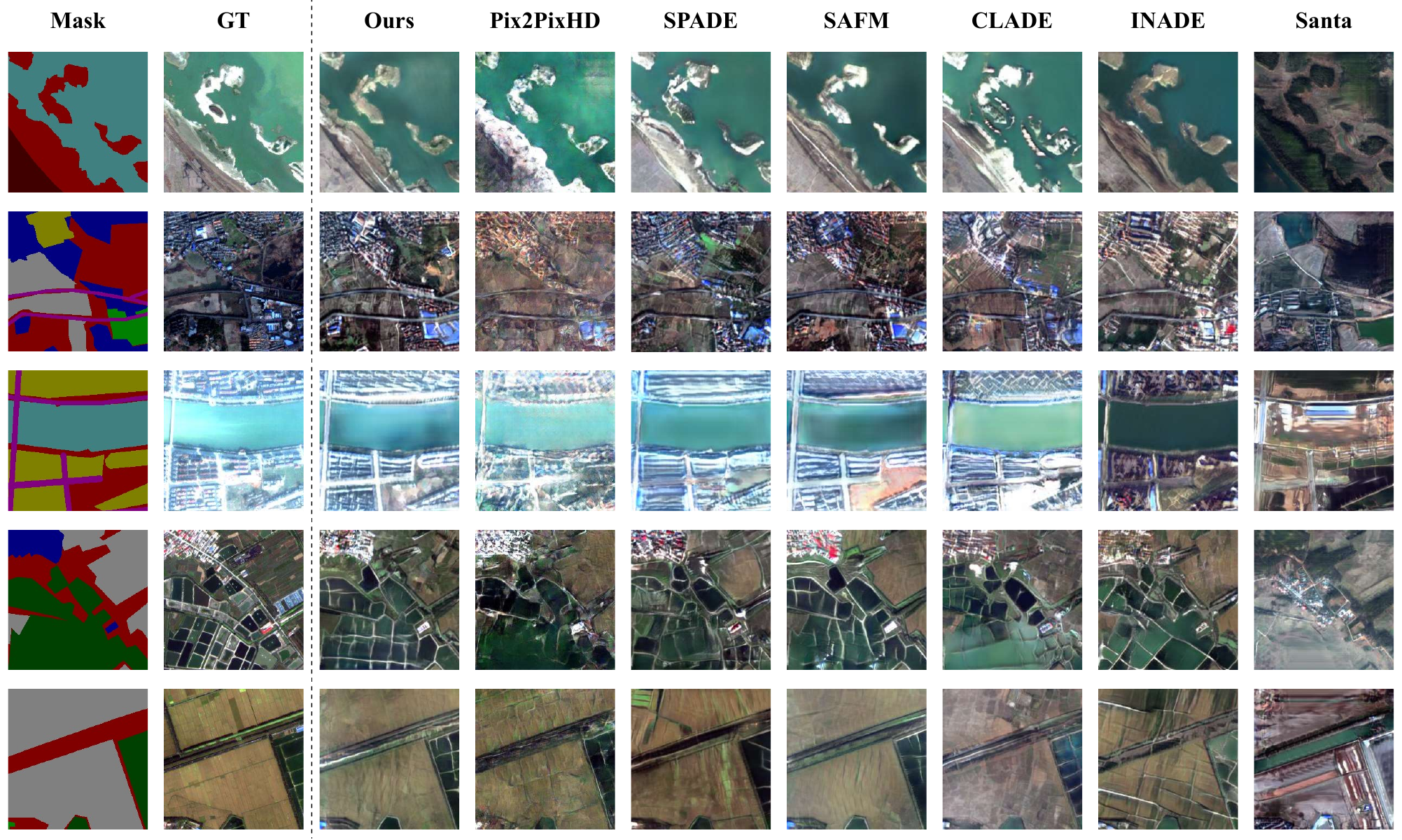}%
\hfil
\caption{Qualitative visualization of synthesis results for various methods on the GID-15 task. The left two columns are the semantic mask and the corresponding ground truth. The other seven columns are the synthesized results. This illustration demonstrates the superiority of our approach in terms of semantic controllability and fine-grained synthesis.}
\label{fig-qualitative}}
\end{figure*}
We compared the images synthesized by our method with those generated by other methods. Fig.~\ref{fig-qualitative} presents a qualitative comparison of our method with other methods on the GID-15 dataset, from which we can observe that:
(1) Our method synthesizes more realistic details (e.g., the farmland section in row 4), due to the utilization of part-level semantic features of remote sensing objects.
(2) Rows 1 and 5 demonstrate that our method effectively handles instances of the same class with varying shapes, whereas other methods fail to produce reasonable results.
(3) Within the framework of the semantic refinement network, our method exhibits strong performance in multi-class, complex and unstructured textures, resulting in improved visual quality, as shown in row 2.
(4) Generative networks that rely solely on simple convolutional operations, without deep feature modulation, often produce significant distortions in the synthesized results, which are heavily affected by contaminated content. For example, in row 3, the outputs of SPADE~\cite{park2019spade} and SAFM~\cite{lv2022safm} appear as long strips, the houses generated by Pix2PixHD~\cite{wang2018p2phd} are fragmented into blocks of pixels, the bottom half of CLADE~\cite{tan2022clade} displays incorrect highlights, and the results of INADE~\cite{tan2023inade} exhibit significant underexposure. In contrast, our model demonstrates greater accuracy and control over the details of the synthesized images through local-level semantic layouts captured by GSD. The resulting images show improved visual consistency and better alignment with human intuition.
\par
In conclusion, our method is more suitable for image synthesis of remote sensing scenes, generating realistic and controllable remote sensing images.
\subsection{Ablation Study}
The ablation experiments cover the following aspects: (1) The contribution of each proposed module to image synthesis results, focusing on semantic controllability, diversity and extensibility. (2) The impact of the loss function on the proposed method's performance. (3) The effect of hyperparameters on model's performance. (4) The robustness of the proposed method in specific scenarios. (5) Some failure cases and their analysis. The baseline for the ablation experiments consists of removing both the SRN and GSD modules, and replacing the HSFM with a simple convolutional layer, as referenced in~\cite{park2019spade}.
\subsubsection{Module Contribution on Controllability}
\label{Contribution_C}
To evaluate the effectiveness of our proposed method in terms of synthesis controllability, we conducted experiments. The details are presented in Table~\ref{table-control}. We evaluated the impact of the GSD, HSFM and SRN modules on model performance to assess controllability. First, removing the GSD module causes a significant drop in synthesized image quality, hindering the model's ability to accurately represent the spatial distribution of remote sensing objects. Next, the absence of the SRN module prevents the model from obtaining fine semantic feedback, further diminishing controllability. The absence of the HSFM module reduces the model's ability to process complex semantic features, limiting its understanding of fine-grained semantics. The results confirm that all proposed modules are effective, and the combination of the GSD, SRN and HSFM modules is crucial for enhancing the model's semantic controllability.

\subsubsection{Module Contribution on Diversity}
\label{Contribution_D}
We evaluated the contribution of the GSD, HSFM and SRN modules to the diversity of synthesized images, as shown in Table~\ref{table-diversity}. Removing the GSD module led to a significant reduction in image diversity, with the model generating more homogeneous styles. Without the SRN module, the model’s ability to generate diverse images under varying semantic conditions (mCSD, mOCD) is limited, and LPIPS increases slightly due to the significant compromise in controllability. Removing the HSFM module clearly limits the model's ability to capture feature variations, resulting in less diverse output images. These experiments clearly demonstrate that the GSD, SRN and HSFM modules play a crucial role in maintaining and enhancing image diversity.
\begin{table}[!t]
    \caption{Individual contribution of proposed modules on Controllability of synthesized results on GID-15 task. The optimal results are highlighted in bold.\label{table-control}}
    \centering
    \begin{tabular}{c|ccc|ccc}
        \toprule
        Method  & SRN & HSFM & GSD  & FID↓ & mIoU↑ & acc↑ \\
        \midrule
        Baseline &  &  &  &  141.408 & 60.471 & 83.067 \\
        \midrule
        +SRN & \checkmark & \checkmark  &  &135.828 & 67.812 & 87.011  \\
        +HSFM  &  & \checkmark & \checkmark & 138.616 & 64.128 & 84.123 \\
        +GSD & \checkmark &   & \checkmark  & 134.071 & 68.094 & 87.785 \\
        \midrule
        +All  & \checkmark & \checkmark & \checkmark  & \textbf{132.283} & \textbf{71.234} & \textbf{88.022} \\
        \bottomrule
    \end{tabular}
\end{table}
\begin{table}[!t]
    \caption{Individual contribution of proposed modules on Diversity of synthesized results on GID-15 task. The optimal results are highlighted in bold.\label{table-diversity}}
    \centering
    \begin{tabular}{c|ccc|ccc}
        \toprule
        Method  & SRN & HSFM & GSD  & LPIPS↑ &  mCSD↑ & mOCD↓ \\
        \midrule
        Baseline &  &  &  & 0.155 & 0.018 & 0.142 \\
        \midrule
        +SRN & \checkmark & \checkmark  &  & 0.184 & 0.021 & 0.076 \\
        +HSFM  &  & \checkmark & \checkmark & \textbf{0.218} & 0.022 & 0.620\\
        +GSD & \checkmark &   & \checkmark & 0.178 & 0.025 & 0.139\\
        \midrule
        +All  & \checkmark & \checkmark & \checkmark  & 0.189 & \textbf{0.028} & \textbf{0.073}\\
        \bottomrule
    \end{tabular}
\end{table}

\subsubsection{Module Contribution on Downstream Tasks}
\begin{figure}[!t]
\centering
\setlength{\tabcolsep}{1mm}{
\includegraphics[width=3.54in]{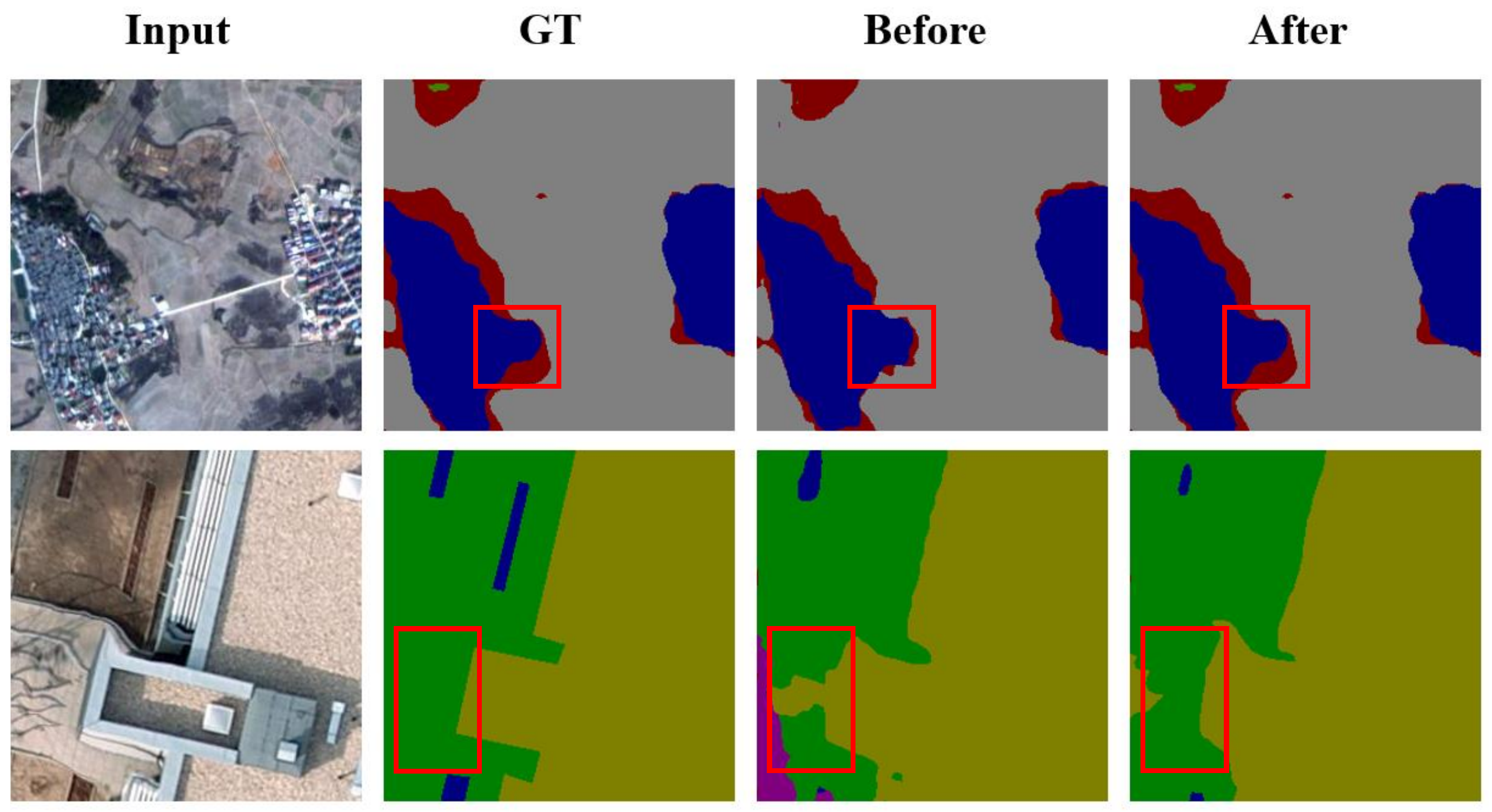}
\caption{Downstream task enhanced performance with results from GID-15 (row 1) and ISPRS (row 2). The left two columns are the images and the corresponding ground truth. The right two columns are the segmentation results before and after enhancement by synthesized images.}
\label{fig-downstream}}
\end{figure}

To reveal the contribution of each module in the proposed method on downstream tasks, we respectively disclose the corresponding modules in Table~\ref{table-downstream}. For brevity, this table presents only absolute values, in contrast to Table~\ref{table-comparison3}. Generally, controllability and diversity of synthesized images are often at odds when used for downstream tasks, such as image semantic segmentation. This creates stringent requirements for the model's overall performance. As discussed in Sec.~\ref{Contribution_C} and Sec.~\ref{Contribution_D}, the proposed GSD and HFSM enhance both controllability and diversity. The SRN sacrifices some diversity but improves fine-grained controllability. The experiments in Table~\ref{table-downstream} support this and confirm our intuition. Fig.~\ref{fig-downstream} compares the segmentation results before and after data enhancement. The results show that the semantic segmentation model performs better with synthetic images as data augmentation.

\begin{table}[!t]
    \caption{Individual contribution of proposed modules in terms of downstream task on GID-15 task. Source only indicates no synthesized images used, and Baseline is the benchmark for removing the three modules. The optimal results are highlighted in bold.\label{table-downstream}}
    \centering
    \setlength{\tabcolsep}{1.5mm}{
    \begin{tabular}{c|ccc|ccc}
        \toprule
        Method  & SRN & HSFM & GSD  & d-mIOU$\uparrow$ & d-FWIoU$\uparrow$ & d- acc$\uparrow$ \\
        \midrule
        Source only &  &  &  & 56.867  & 68.250  & 80.814  \\
        \midrule
        Baseline &  &  &  & 58.074  & 68.941  & 81.362  \\
        +SRN & \checkmark & \checkmark  &   & 58.738  & 69.271  & 81.852  \\
        +HSFM  &  & \checkmark &  \checkmark & 58.553  & 69.137  & 81.608  \\
        +GSD &  \checkmark &   & \checkmark & 58.879  & 69.424  & 81.669  \\
        \midrule
        +All  & \checkmark & \checkmark & \checkmark  & \textbf{59.191} & \textbf{69.747} & \textbf{82.002} \\
        \bottomrule
    \end{tabular}}
\end{table}

\subsubsection{Impact of Loss Function}
To further investigate the effect of the proposed loss function on the results, we conducted a loss function ablation experiment. In this experiment, we used different training losses: baseline, using only $L_{ref}^I$, combining $L_{ref}^I$ and $L_{ref}^z$, combining $L_{ref}^z$ and $L_{ref}^S$ and combining all. Where Baseline stands for using only $L_{adv}$, $L_{fm}$ and $L_{perc}$, which are widely used in semantic image synthesis.
Table~\ref{table-loss} shows a significant improvement in the FID metrics after incorporating all terms of $L_{ref}$. This improvement is due to $L_{ref}$ providing fine-grained semantic feedback and enhanced controllability to the synthesized images, resulting in more realistic remote sensing images with improved semantic control. The baseline uses only the three loss functions mentioned earlier, resulting in inferior performance. $L_{ref}^I$ and $L_{ref}^z$, which use segmentation results for semantic feedback, fail to account for errors from the SRN network. Although combining $L_{ref}^I$ and $L_{ref}^S$ alleviates some of these issues, it still lacks direct semantic feedback. The inclusion of all components allows the generative network to integrate fine-grained semantic feedback and provide pixel-level compensation, optimizing the model to better align with human perception, resulting in a 3.895 improvement in FID.

\begin{table}[!t]
    \caption{Impact of loss function on GID-15 task. The optimal results are highlighted in bold. \label{table-loss}}
    \centering
    \begin{tabular}{c|ccc|c}
        \toprule
        Method  & $L_{ref}^I$ & $L_{ref}^z$ & $L_{ref}^S$  & FID \\
        \midrule
        Baseline &  &  &  &  136.178 \\
        \midrule
        +$L_{ref}^I$ & \checkmark &  &   & 136.071 \\
        +$L_{ref}^I$  & \checkmark & \checkmark &   & 134.296 \\
        +$L_{ref}^z$  & \checkmark &  & \checkmark  & 135.741 \\
        \midrule
        +All  & \checkmark & \checkmark & \checkmark  & \textbf{132.283} \\
        \bottomrule
    \end{tabular}
\end{table}

\subsubsection{Impact of Hyper-parameters}
\label{Sec-ablation-hyper}
There are several hyperparameters in the proposed network, including $n_\theta$ and $n_\rho$ introduced in Sec.~\ref{Sec-GSD} and $\Gamma$ in Sec.~\ref{Sec-Lref}. This section explores the effect of parameter settings on final performance. As is shown in Table~\ref{table-Epoch} and Table~\ref{table-mn}. Table~\ref{table-Epoch} illustrates the epoch at which semantic feedback starts to affect model performance. Table~\ref{table-mn} demonstrates the influence of $n_\theta$ and $n_\rho$ on GSD feature representation and their impact on performance. Note that these hyperparameters are not designed for a specific dataset, although we only present the quantitative results for GID-15.
\begin{table}[!t]
    \caption{Impact of the epoch when SRN gives semantic feedback on GID-15 task. The optimal results are highlighted in bold.}
    \centering
    \setlength{\tabcolsep}{1mm}{
    \begin{tabular}{c|ccccccc}
        \toprule
        Epoch & 0 & 50 & 60 & 70 & 80 & 100 & 150 \\
        \midrule
        FID↓ & 141.432 & 137.791 & 136.851 & 134.294 & \textbf{132.283} & 133.103 & 135.587\\
        \bottomrule
    \end{tabular}}
    \label{table-Epoch}
\end{table}

\begin{table}[!t]
    \caption{Impact of the values of $n_\theta$ and $n_\rho$ in proposed GSD method on GID-15 task. The optimal results are highlighted in bold. \label{table-mn}}
    \centering
    \begin{tabular}{c|cccc}
        \toprule
        ($n_\theta$,$n_\rho$)  & (3,24) & (6,12) & (12,6) & (24,3) \\
        \midrule
        FID↓ & 141.755 & 135.146 & \textbf{132.283} & 138.791 \\
        \bottomrule
    \end{tabular}
\end{table}

\subsubsection{Robustness to the Unknown Class}
\begin{figure}[!t]
\centering
\setlength{\tabcolsep}{1mm}{
\includegraphics[width=3.5in]{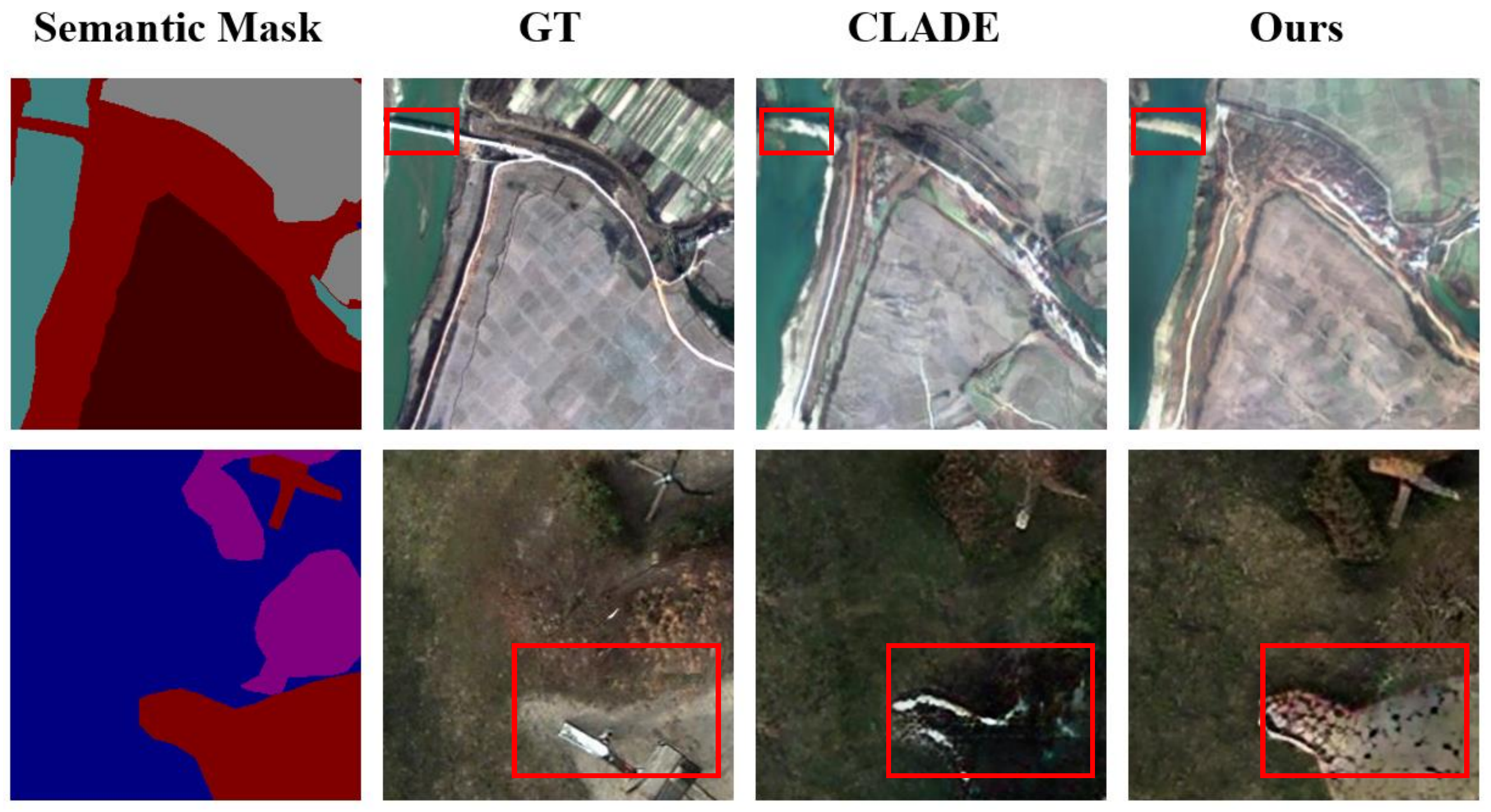}
\caption{Robustness to the unknown class, the results are from both GID-15 (row 1) and ISPRS (row 2). The left two columns are the semantic mask and the corresponding ground truth. The right two columns are the synthesized results.}
\label{fig-unknow}}
\end{figure}

In image synthesis tasks, background categories are often excluded due to their complex details, variations, and contextual dependencies with neighboring semantic classes. However, background classes often make up a significant portion of remote sensing images. This section reveals the remarkable robustness of the proposed method to unknown background categories. As shown in Fig.~\ref{fig-unknow}. Other methods~\cite{tan2022clade} that synthesize the background class (light red) produce issues such as foreground-background confusion. By incorporating part-level semantic features, HySEGGAN accurately captures the complex geometric features and diversity of the background, maintaining flexibility in handling changing natural backgrounds. Additionally, our approach fully accounts for the contextual relationship between the background and foreground, ensuring that the synthesized background is visually coherent with neighboring foreground objects and minimizing the appearance of blurring or artifacts.
\subsubsection{Failure Example and Analysis}
\begin{figure}[!t]
\centering
\setlength{\tabcolsep}{1mm}{
\includegraphics[width=3.54in]{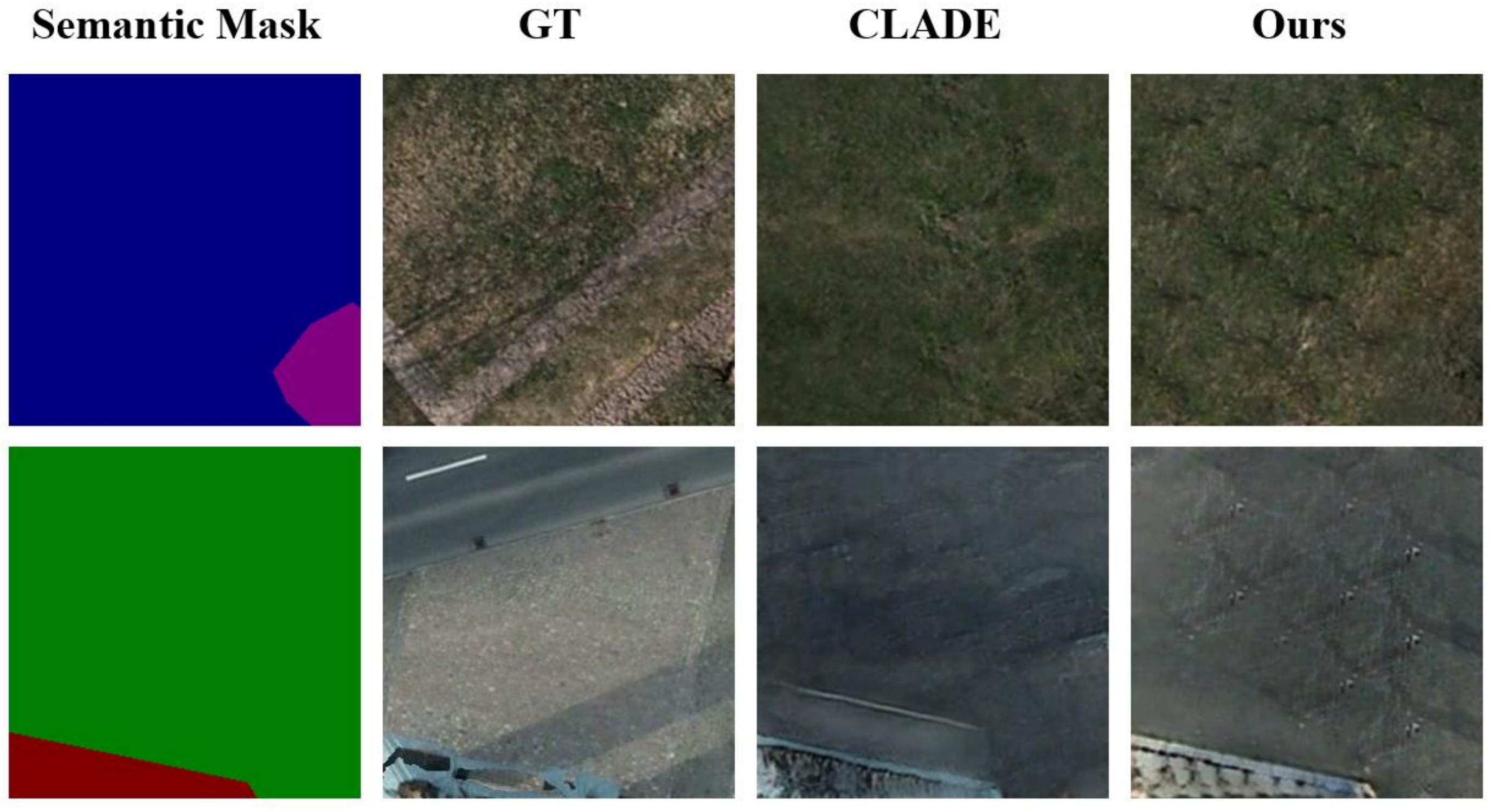}
\caption{Failure examples of the proposed method, the results are from ISPRS task. The left two columns are the semantic mask and the corresponding ground truth. The right two columns are the synthesized results.}
\label{fig-failure}}
\end{figure}

The above experiments validate the efficiency of the model in the benchmark. However, we found that our model has limited performance in cases where a particular semantic class in the input mask is overly dominant. For example, in Fig.~\ref{fig-failure}, large areas of pavement or shrubs cause certain features or textures to repeat throughout the synthesized image, leading to pattern collapse. Other models also exhibit similar issues, which may represent an important challenge to address in future semantic image synthesis tasks for remote sensing images.

\section{Conclusion}
In this paper, we address the challenge of semantic image synthesis for remote sensing images by exploring novel aspects of semantic controllability, diversity and extensibility. A geometric information characterization method is proposed for part-level semantic representations. The hybrid semantic feature modulation block we employ allows for precise alignment between synthesized images and masks, integrating both global and local semantic information effectively. Furthermore, a novel loss function is introduced to mitigate the semantic confusion problem, enhancing the model's robustness and supporting reliable content synthesis. Comprehensive evaluations demonstrate that HySEGGAN delivers superior visual quality and improves the performance of downstream tasks. Particularly, the proposed model is highly efficient in challenging scenarios, such as the synthesis of the background class. Overall, our approach establishes a new benchmark for the synthesis of remote sensing images.
\par
However, the limitations of the proposed model mainly lie in dominance confusion, where the absolutely dominant class may experience semantic confusion alongside neighboring classes. In future work, we aim to further investigate the interplay between feature description engineering and image synthesis models.

\normalem
	\bibliographystyle{IEEEtran}
	\bibliography{refs}

\vfill

\end{document}